# NL-CNN: A Resources-Constrained Deep Learning Model based on Nonlinear Convolution

Radu Dogaru and Ioana Dogaru,
Polytechnic University of Bucharest, Dept. of Applied Electronics and Information Engineering
radu.dogaru@upb.ro; ioana.dogaru@upb.ro

*Abstract* — A novel convolution neural network model, abbreviated NL-CNN is proposed, where nonlinear convolution is emulated in a cascade of convolution + nonlinearity layers. The code for its implementation and some trained models are made publicly available. Performance evaluation for several widely known datasets is provided, showing several relevant features: i) for small / medium input image sizes the proposed network gives very good testing accuracy, given a low implementation complexity and model size; ii) compares favorably with other widely known resources-constrained models, for instance in comparison to MobileNetv2 provides better accuracy with several times less training times and up to ten times less parameters (memory occupied by the model); iii) has a relevant set of hyper-parameters which can be easily and rapidly tuned due to the fast training specific to it. All these features make NL-CNN suitable for IoT, smart sensing, bio-medical portable instrumentation and other applications where artificial intelligence must be deployed in energy-constrained environments.

*Keywords*: convolution neural network, resources-constrained implementation, nonlinear convolution, computer vision

## I. INTRODUCTION

Artificial intelligence applications are now wide spread. In many circumstances, such as internet of things (IoT), smart sensors, mobile applications and so forth one has to achieve a good performance (often measured as classification accuracy) with a low complexity. Proposals for energy-efficient artificial neural network abound in recent years. Some of them focus on hardware implementations while some on the model description. The research presented here falls in the second category. So far, numerous such resources-constrained models were proposed (MobileNet [1], EffNet [2] and more recently MobileExpressNet [3] to name just a few), some of which being already embedded in major AI development frameworks such as Keras / Tensorflow[1]. While such models are widely accepted, in our experience some drawbacks were noted: i) Such models are usually designed for large input image sizes, but in many resources-constrained applications some 32x32 image size often suffice to get a good accuracy while significantly improving the efficiency (knowing that convolution over large image sizes is computationally intensive operation); ii) Their structure is rather clumsy, with a diversity of layers types, consequently such a complicated structure may lead to various problems, particularly when hardware implementation is desired. In any case, we observed relatively large training times, most likely as a consequence of their clumsy internal structure; iii) Usually, models such as mentioned above are often pre-defined, with a very small number of freely tunable hyper-parameters (e.g. only the *alpha* parameter in MobileNet). Consequently, inspired from the work in [4] where it is shown that with a careful design of the number of filters in each convolution layer a good performance can be achieved, we propose herein NL-CNN, a relatively compact and easy to understand structure, where relatively fast training ensured near state-of-the-art performance for a wide variety of datasets. Fast training ensures the capability to rapidly optimize a set of at least 4 hyper-parameters such that compactness and good accuracy is achieved for each particular dataset considered.

Recently in [5] we proposed a compact solution for signal classification where a specially designed feature extractor converted signals into image spectrograms then submitted to a specially designed convolution neural network, denoted here NL-CNN, (NL standing from NonLinear) with 3 macro-layers. In each macro-layer non-linear convolution (NL) [6][7] is emulated using a cascade of *nl* readily available Keras/Tensorflow layers (each level having a traditional (linear) convolution followed by a ReLu nonlinearity). This paper expands the research on NL-CNN with an improved, publicly available model [8] and investigates its capabilities for a wider spectrum of datasets in order to compare its performance to state of the art solutions using other resources-constrained models. The result is a low complexity (evaluated in terms of parameters and computing latency) classifier, yet capable to provide state of the art accuracy. Consequently, it replaces conveniently more traditional models for resource-constrained CNNs, for instance, as shown in section IV, we found that it performs better than MobileNetv2 for all of the datasets investigated herein (using small/medium input image sizes).

Section II describes the NL-CNN model emphasizing on the specific hyper-parameters, Section III indicates how hyper-parameters can be optimized for best performance, while Section IV presents NL-CNN performance evaluated for some significant and widely used datasets in computer vision. Concluding remarks are given in Section V.

## II. NL-CNN MODEL AND ITS IMPLEMENTATION

The key ingredient of the NL-CNN is a "macro-layer" with up to *nl* convolution layers emulating nonlinear convolution, as shown in Fig.1. The function defining the NL-CNN model is available in [8]. It is callable as:

---
[1] https://keras.io/

```
model = create_nl_cnn_model(input_shape, num_classes, k=1.5,
separ=0, flat=0, width=80, nl=(2,2), add_layer=1),
```

with specific parameters explained in Fig.2 and in the remaining part of the paper. While nonlinear convolution was found already as a good solution to improve performance in computer vision [7] and other areas [6], it is here emulated by cascading *nl* convolution layers with the same *nf* number of filters, each followed by a nonlinearity (e.g. ReLU). For a given *nl*, each pixel in the *nf* filters of the macro-layer outputs is calculated as a nonlinear convolution with respect to a $3+2(nl-1)$ x $3+2(nl-1)$ neighborhood from the input image (assuming 3x3 convolution kernels in each convolution layers). As seen in Fig.1 the output nonlinearity is achieved into a max-pool layer operating in 4x4 neighborhoods and with a stride of 2, ensuring 2 times image size reduction after each macro-layer.

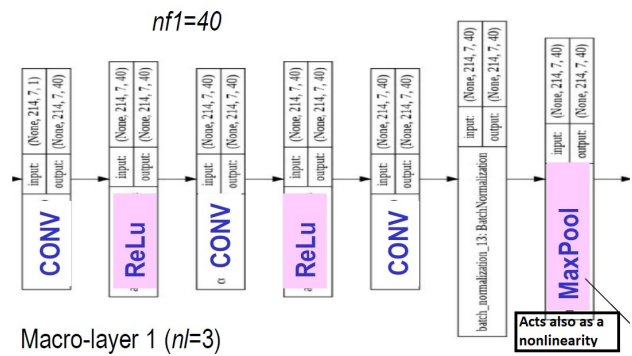

Figure 1: The structure of one macro-layer in the NL-CNN, emulating nonlinear-convolution with a given number of filters (here *nf1=40*).

The entire network stacks 3 or 4 such macro-layers (the 4'th is usually considered if larger input image sizes are considered). The description of the entire NL-CNN and its associated hyper-parameters is given in Fig. 2.

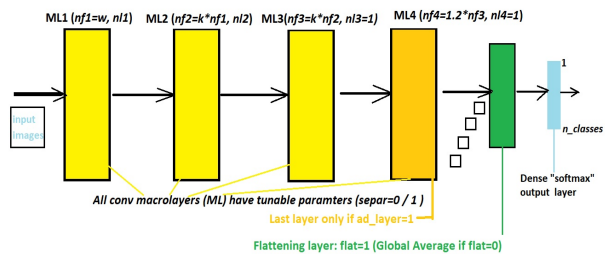

Figure 2: The structure of the NL-CNN network and its hyper-parameters.

### III. OPTIMIZING THE HYPER-PARAMETERS

In order to get the best accuracy from the NL-CNN network one has to carefully tune its hyper-parameters. The following examples are given for the widely known MNIST dataset [9] while a similar process is considered for any other dataset. The MNIST set consists in 60000 gray-level (one-channel) training images and 10000 for testing. The dimensions of images are 28x28. In order to have a faster training, a limited number of epochs (here 20) and a relatively large batch size (here 500) is chosen. The most critical parameters are: *nl* (the nonlinearities in the first and second macro-layers); *k* (a multiplying factor for the number of filters in each layer); and *w* (width – number of convolution filters or channels for the input macro-layer). The parameters controlling the NL-CNN size are set to: *separ=0* (normal convolution) and *flat=1* (normal flattening layer).

Next table considers some results for ***nl tuning*** (while *k=2*, *w=20*): As seen, the best choice is *nl=(2,2)*. As in any other over-fitting case, using larger values would not increase accuracy at the cost of more parameters.

TABLE I. TUNING THE *NL* PAIR HYPER-PARAMTERES FOR MNIST

| nl | (1,1) | (2,1) | (3,1) | (4,1) | (1,2) | (2,2) | (3,2) |
|---|---|---|---|---|---|---|---|
| Accuracy (%) | 99.11 | 99.04 | 99.15 | 99.08 | 99.33 | 99.41 | 99.30 |
| Parms | 49690 | 53310 | 56930 | 60550 | 64130 | 67750 | 71370 |
| Train time | 36 | 45 | 53 | 61 | 43 | 52 | 60 |

Next, after choosing the best case i.e. *nl=(2,2)* next hyper-parameter to tune is the width *w*. Usually is good to start with a small value and increase it slightly until some saturation is observed in the accuracy parameter. It was observed that for larger *w* a good compromise between accuracy and size is obtained while choosing the *flat=0* (global average flattening layer). As seen, using *flat=0* ensures a better accuracy using less model parameters. As in the previous case, an optimal value for *w* was found; larger values do not increase but rather decrease the test accuracy.

TABLE II. TUNING OF *W* PARAMETER FOR MNIST

| W | 10 | 15 | 20 | 25 | 30 | 40 |
|---|---|---|---|---|---|---|
| Accuracy (%) | 98.95 | 99.08 | 99.41 | 99.37 | 99.49 | |
| Parms | 20380 | 40690 | 67750 | 101560 | 142120 | |
| Train time | 36 | 42 | 52 | 57 | | |
| Flat=0 (Global average) | | | | | | |
| Accuracy (%) | | | | | 99.57 | 99.55 |
| Parms | | | | | 124120 | 219490 |
| Train time | | | | | 62 | 82 |

Finally, although our experience with a wide variety of datasets has shown that *k=2* is good option, it is useful to try various other values (usually for smaller values getting more compact networks, with some loss in performance, as seen in the next table).

TABLE III. TUNING THE EXPANSION PARAMETER *K* FOR MNIST

| k | 1.25 | 1.5 | 1.75 | 2 | 2.25 | 2.5 |
|---|---|---|---|---|---|---|
| Accuracy (%) | 99.29 | 99.47 | 9949 | 99.57 | 99.50 | 99.55 |
| Parms | 47101 | 67345 | 91201 | 124120 | 160153 | 208915 |
| Train time | 56 | 57 | | 62 | 69 | 72.3 |

As seen, in this example, the entire tuning process lasted about 18 minutes, as a consequence of the relatively fast training (when compared to other compact models such as MobileNet – details in next section). Finally, with the chosen set of hyper-parameters *(k=2; w=30; nl=(2,2); separ=0; flat=0; add_layer=0)* one should try a longer training process (more epochs, here 100) with some smaller batch size. In the case of MNIST this results in a model with 99.7 % accuracy. No image augmentation technique was used for this and for all other datasets considered in this paper. The model is stored in [8]. This performance is quite good even if compared with state of the art (about 99.9%) given the compactness of the resulted model (as seen in Table VI, MobileNetV2 from Keras achieves only 99.49% with a model occupying more than 10 times more memory)

Several other implementations of MobileNet can be considered. The above mentioned Keras model required to increase the image size and add 2 channels in order to fit the requirements of given implementation. An implementation from [10] is also used with various *alpha* values. The result for MNIST in the same training context (batch=100, epochs=100, Adam optimizer) indicates larger training times, and more parameters than in the NL-CNN to achieve a relatively low accuracy on the test set:

TABLE IV. PERFORMANCES OBTAINED WITH A MOBILE-NET FROM [10]

| alpha | 0.75 | 1 |
|---|---|---|
| Accuracy(%) | 99.02 | 99.04 |
| Train time (s) | 548 | 713 |
| Parameters | 1.807.354 | 3.181.898 |

In order to reduce the size of the network one may use the "separable-conv2d" instead of the "classic" convolution by choosing *separ=1*. Similar hyper-parameter optimization should be performed, but now it is better to use larger *w*. Next table shows indeed that very compact models can be achieved at the expense of lower accuracies:

TABLE V. TUNING THE *NL* FOR THE COMPACT VERSION OF NL-CNN

| nl | (1,1) | (2,1) | (3,1) | (4,1) |
|---|---|---|---|---|
| Accuracy (%) | 95.28 | **97.31** | 89.75 | 32.48 |
| Parms | 18079 | 18679 | 19279 | 19879 |
| Train time | 35 | 43 | 50 | 58 |

## IV. PERFORMANCE ON VARIOUS DATASETS

The following table gives a synthesis with the best solutions (models) obtained using NL-CNN for a wide variety of datasets. References to the specific sets are given in the same table, with best known performance (although not on a compact network, most taken from https://benchmarks.ai/) given in parenthesis. Models with mention "light" are usually obtained by choosing the parameter *separ=1*. For each line in the table, the .h5 model is available in [8] (It can be located based on model type and accuracy: for instance **nlcnn_mnist_99_72.h5** corresponds to the NL-CNN trained with MNIST dataset and achieving 99.72% accuracy). For comparison with other, widely used, compact models, the same datasets were used for training the pre-trained MobileNetv2 network with *alpha=0.75* (code included in [8]) and the resulting accuracies and model files sizes are reported in the next table. The same training conditions were considered in both cases, i.e. 100 epochs with batches sizes depending on the set, usually 100.

TABLE VI. PERFORMANCE OF THE NL-CNN AND MOBILENET MODELS FOR VARIOUS DATASESTS

| DATASET Info about NL-CNN | Accuracy (%) | .h5 model Mbytes | MobileV2 Acc (%) | .h5 model Mbytes | Relevant refs. |
|---|---|---|---|---|---|
| **MNIST** | **99.7** | 1.57 | 99.49 | 17.2 | [9] 99.8% |
| **EMNIST** balanced | **90.47** | 5.64 | 87.69 | 17.7 | [11] 90.59% |
| **SVHN** | **96.15** | 4.2 | | | [12] 99% |
| **CIFAR10** Improved Light | 88.60 **90.69** 86.08 | 5.4 6.02 0.85 | 80.03 | 17.2 | [13] 99.37% |
| **GTRSB** Light | **98.97** 98.84 | 2.974 0.7 | 96.13 | 17 | [14][2] 99.46% |
| **FRUITS**[3] | **99.12** | 0.69 | 99 | 1.53 | [15] 96.7% |
| **FER2013** Light 4-layers | 63.86 61.52 **67.62** | 4.86 0.52 6.08 | 58.56 | 17.1 | [16] 75.2% [3] 67% compact model |
| **F-MNIST** | **93.62** | 4.87 | 92.73 | 17.2 | [17][4] 96.91 |

In all cases, the number of parameters is about 12 times less than the number of bytes represented by the .h5 files. Note than for all datasets with only one image channel (MNIST, FER2013, F-MNIST), some additional processing was considered to create 3-channels datasets with 32x32 size resolutions (to keep compatibility with MobileNetV2 models available in Keras). Notably, in all circumstances the training times for the MobileNet models are 3-4 times larger (8-15 seconds per epoch) than for the optimal NL-CNN (2-5 seconds per epoch). These results clearly demonstrate that properly tuned NL-CNN models compete favorably in both performance and compactness with widely accepted models such as MobileNet_v2.

Moreover, additional careful optimization of hyper-parameters may improve the results presented in the above

---
[2] https://benchmark.ini.rub.de/gtsrb_news.html
[3] Image size down-sampled to 32x32, the 120 classes version
[4] https://paperswithcode.com/sota/image-classification-on-fashion-mnist

table. For instance, in the case of the emotion faces FER2013 dataset, adding an extra-layer (4-th macr-olayer) with properly tuning of *k* and *w* parameters improves the testing accuracy to 67.62%, quite close to the best result reported so far (75.3%) on a non-compact architecture [16]. Recently a compact model proposed in [3] reported 67% on the same dataset.

## V. CONCLUDING REMARKS

A novel compact yet accurate convolution neural model is proposed, implementation code and some trained models are made available in [8]. Particularly for small or medium image sizes (up to 128x128), NL-CNN provides a simple and efficient solution, with relatively fast training allowing a fine optimization of the hyper-parameters. Accuracies are in the range of values obtained with state-of-the art solutions and in some cases larger than what is reported so far. Our model compares favorably to well established resources-constrained MobileNet_v2 model, with faster training (around 3 times) better test accuracy and up to 10 times less complexity (measured as memory occupied by the .h5 model file).

Resuming, the following are relevant features of the proposed architecture: i) Quite accurate, given the reduced complexity, compares favorably to MobileNet; ii) Relatively simple to understand (and implement, in HW oriented solutions such as FPGA); iii) Training speed is better than in MobileNet thus allowing for a rapid and careful optimization of the hyper-parameters for the best performance; iv) The code is readily available [8] as notebook (tested in Google COLAB[5]). Further research may focus on datasets with larger images, solutions for FPGA implementations including aggressive quantization of parameters and some architecture improvements.

---

[5] https://colab.research.google.com/